\documentclass[sigconf,nonacm]{aamas} 

\usepackage{balance} 
\usepackage{threeparttable}
\usepackage{graphicx}
\usepackage{caption}
\usepackage{subcaption}
\usepackage{float}

\setcopyright{ifaamas}
\acmConference[AAMAS '23]{Proc.\@ of the 22nd International Conference
on Autonomous Agents and Multiagent Systems (AAMAS 2023)}{May 29 -- June 2, 2023}
{London, United Kingdom}{A.~Ricci, W.~Yeoh, N.~Agmon, B.~An (eds.)}
\copyrightyear{2023}
\acmYear{2023}
\acmDOI{}
\acmPrice{}
\acmISBN{}

\title[]{Benchmarking Robustness and Generalization in Multi-Agent Systems: A Case Study on Neural MMO}

\author{Yangkun Chen\textsuperscript{*1,3}}
\affiliation{
  }
\email{chen-yk21@mails.tsinghua.edu.cn}

\author{Joseph Suarez\textsuperscript{*\dag2}}
\affiliation{
  }
\email{jsuarez@mit.edu}

\author{Junjie Zhang\textsuperscript{*1,3}}
\affiliation{
  }
\email{zhangjj21@mails.tsinghua.edu.cn}

\author{Chenghui Yu\textsuperscript{1,3}}
\affiliation{
  }
\email{ych20@mails.tsinghua.edu.cn}

\author{Bo Wu\textsuperscript{3}}
\affiliation{
  }
\email{bowu@chaocanshu.ai}

\author{Hanmo Chen\textsuperscript{1,3}}
\affiliation{
  }
\email{chm20@mails.tsinghua.edu.cn}

\author{Hengman Zhu\textsuperscript{3}}
\affiliation{
  }
\email{henryzhu@chaocanshu.ai}

\author{Rui Du\textsuperscript{4}}
\affiliation{
  }
\email{durui@bilibili.com}

\author{Shanliang Qian\textsuperscript{4}}
\affiliation{
  }
\email{qianshanliang@bilibili.com}

\author{Shuai Liu\textsuperscript{4}}
\affiliation{
  }
\email{liushuai01@bilibili.com}

\author{Weijun Hong\textsuperscript{5}}
\affiliation{
  }
\email{hongweijun@corp.netease.com}

\author{Jinke He\textsuperscript{6}}
\affiliation{
  }
\email{J.He-4@tudelft.nl}

\author{Yibing Zhang\textsuperscript{7}}
\affiliation{
  }
\email{554011619@qq.com}

\author{Liang Zhao\textsuperscript{8}}
\affiliation{
  }
\email{zhaoliang@idea.edu.cn}

\author{Clare Zhu\textsuperscript{9}}
\affiliation{
  }
\email{czhu529@gmail.com}

\author{Julian Togelius\textsuperscript{10}}
\affiliation{
  }
\email{julian.togelius@nyu.edu}

\author{Sharada Mohanty\textsuperscript{11}}
\affiliation{
  }
\email{mohanty@aicrowd.com}

\author{Jiaxin Chen\textsuperscript{\dag 3}}
\affiliation{
  }
\email{jiaxinchen@chaocanshu.ai}

\author{Xiu Li\textsuperscript{\dag 1}}
\affiliation{
  }
\email{li.xiu@sz.tsinghua.edu.cn}

\author{Xiaolong Zhu\textsuperscript{\dag 3}}
\affiliation{
  }
\email{xiaolongzhu@chaocanshu.ai}

\author{Phillip Isola\textsuperscript{2}}
\affiliation{
  }
\email{phillipi@mit.edu}

\newcommand\blfootnote[1]{%
  \begingroup
  \renewcommand\thefootnote{}\footnote{#1}%
  \addtocounter{footnote}{-1}%
  \endgroup
}
\begin{abstract}
We present the results of the second Neural MMO challenge, hosted at IJCAI 2022, which received 1600+ submissions. This competition targets robustness and generalization in multi-agent systems: participants train teams of agents to complete a multi-task objective against opponents not seen during training. The competition combines relatively complex environment design with large numbers of agents in the environment. The top submissions demonstrate strong success on this task using mostly standard reinforcement learning (RL) methods combined with domain-specific engineering. We summarize the competition design and results and suggest that, as an academic community, competitions may be a powerful approach to solving hard problems and establishing a solid benchmark for algorithms. We will open-source our benchmark including the environment wrapper, baselines, a visualization tool, and selected policies for further research. \blfootnote{*: Equal contribution. \dag: Corrosponding author. 1: Shenzhen International Graduate School, Tsinghua University. 2: Massachusetts Institute of Technology. 3: Parametrix.ai. 4: bilibili. 5: NetEase Games AI Lab. 6: Delft University of Technology. 7: Chengdu Goldwin Electronics Technology Co., Ltd. 8: International Digital Economy Academy. 9: Stanford University. 10: New York University. 11: AICrowd. Correspondence to Jiaxin Chen (Parametrix.ai), Xiu Li(Shenzhen International Graduate School, Tsinghua University), Joseph Suarez (MIT), and Sharada Mohanty (AICrowd).}
\end{abstract}


\newcommand{\BibTeX}{\rm B\kern-.05em{\sc i\kern-.025em b}\kern-.08em\TeX}

\begin{document}


\pagestyle{fancy}
\fancyhead{}


\maketitle

\section{Introduction}

Real-world applications of reinforcement learning (RL) require robust algorithms  \cite{generalization} that can adapt to dynamic environments. While substantially studied in single-agent RL \cite{ProcGen_env,ProcGen_competition}, this subject has been less explored in multi-agent systems. This is of particular importance to multi-agent RL (MARL) algorithms because learned policies must adapt to changes in other agents' behaviors in addition to changes in the environment. Here we suggest three difficulties for establishing benchmarks in multi-agent systems that should resonate with MARL researchers:

\begin{enumerate}
    \item \textbf{Lack of environments:} while there are many single-agent environments of varying complexities that are standard, efficient, and simple to use, few multi-agent environments satisfy all three of these properties.
    \item \textbf{Lack of infrastructure:} most RL libraries and interfaces are intended for single-agent systems, but multi-agent training requires scalability, flexibility, and other additional features. For example, an accurate skill-rating system is needed for multi-agent evaluation as performance is relative to other agents.
    \item \textbf{Lack of domain-specific optimization:} minor implementation details and domain-specific tricks like feature engineering often highly influence the final performance of RL algorithms \cite{implementation}. Although these techniques are not the focus of academic research, without them, it is hard to identify the roots of algorithmic progress and benchmark algorithms fairly.
\end{enumerate}

This paper summarizes the IJCAI 2022 Neural MMO challenge and offers a solution to these three problems. Neural MMO is a good environment to start with because it supports large-scale populations, is computationally efficient and is actively maintained. In addition to the environment, we built a large-scale parallel evaluation tool and a TrueSkill\cite{herbrich2006trueskill} rating system on the AICrowd platform as the infrastructure. The competition among participants provides an inherent incentive for domain-specific optimization, which is often overlooked in academic research. We hope that our methodology can serve as a stepping stone towards establishing more general benchmarks and promoting future research in Neural MMO and other multi-agent systems. Our main contributions are:

\begin{enumerate}
    \item \textbf{Orchestration:} we detail the structure of our competition, including the environment, resources, the design of tracks, and the evaluation system. While RL competitions are gaining popularity, few resources exist on how to design a good competition. We believe this will be useful to guide future RL competitions.
    \item \textbf{Insights:} we analyze the emergent behaviors and strategies over the 1600+ submissions received and provide insights about the dynamics of the unique multi-agent system consisting of different participants. For example, we find an interesting arms race between rule-based methods and learning-based methods.
    \item \textbf{Policy Pool:} we release a pool of 20 submitted policies to promote future research on Neural MMO. The policy pool is diverse, containing both rule-based and RL-based implementations of aggressive and conservative strategies. This will be useful in evaluating policy robustness against a variety of opponents.
\end{enumerate}

\begin{figure*}[htbp]
    \centering
    \includegraphics[width=1\textwidth]{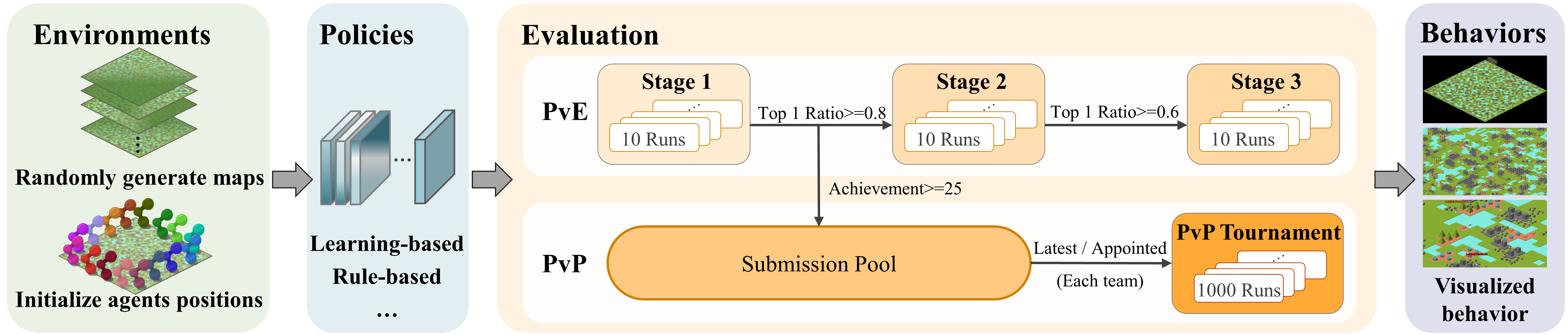}
    \caption{Evaluation structure of the competition. Submitted policies are evaluated in two stages. In the PvE track, there are multiple qualifying rounds against increasingly difficult built-in opponents, and participants receive feedback within minutes of submitting. The PvP track features weekly tournaments to determine the relative skill of all qualified submissions.}
    \label{fig:framework}
\end{figure*}
\section{Related Works}

\subsection{Environments and Benchmarks} 
In recent years, wrapping existing games as environments has been a popular approach to increase complexity and promote novel algorithms. MineRL\cite{MineRL} uses Minecraft to highlight hard problems such as hierarchical task structure and sparse rewards. The NetHack Learning Environment \cite{nethack_env} provides a rich and challenging environment focused on the problems of exploration and skill acquisition while allowing fast simulation. ProcGen \cite{ProcGen_env} uses 16 procedurally-generated gym environments and is designed to benchmark both sample efficiency and generalization in reinforcement learning. 

Among multi-agent environments, the most commonly used is the StarCraft Multi-agent Challenge (SMAC) \cite{smac} from the game StarCraft2. SMAC is mainly intended to investigate algorithms for multi-agent cooperation. Google Research Football (gfootball) \cite{gfootball} uses a physics-based 3D football environment in multi-player and multi-agent scenarios, proposed to benchmark algorithms on the sparse reward and multi-agent cooperation. Neural MMO \cite{nmmo19} proposes an open-ended Massively Multiplayer Online (MMO) environment with up to $1024$ agents to study robustness and teamwork in a massive-agent environment. Agar.io \cite{agar} and the similar GoBigger \cite{gobigger} uses the popular online multi-player game Agar.io\footnote{\textcolor{blue}{\url{https://agar.io/}}} for multi-agent cooperation and competition. We use Neural MMO as our competition environment because it supports large-scale population simulation with up to $16$ teams in one environment. 

\subsection{RL Competitions} 

Several works \cite{benchmark_RL} attempt to establish solid benchmarks for deep RL algorithms. A key issue in doing so is that performance highly depends on minute implementation details \cite{implementation}, which are often not the focus of academic research. One alternative is an open competition that provides a natural incentive for domain-specific optimization: winning. This format has become popular in recent years, and existing competitions can be roughly categorized into three classes: PvE, 1v1, and FFA. 

\textbf{PvE} (Player vs Environment): these competitions evaluate agents against preset (usually randomized) environments as specific algorithmic benchmarks. For example, the NetHack challenge \cite{nethack_2021} concentrates on sparse reward and exploration while the MineRL competition \cite{MineRL_competition} concentrates on sample efficiency. However, in PvE settings, agents are evaluated against given environments or fixed bots instead of other learning agents. This evaluation paradigm limits the ability to benchmark robustness and generalizability to new opponents.

\textbf{1v1} or one team vs another: these competitions evaluate agents against other participants' policies in a two-agent or two-team mode, such as Google Research Football \cite{gfootball} and Lux-AI \cite{Lux_AI_Challenge_S1}. This requires agents to adapt to different kinds of opponents instead of specializing in a fixed environment. 

\textbf{FFA} (Free-for-All): this setting places many independent agents or many teams in the same shared environment. Compared to the 2-team mode, FFA competitions can create a vast space of cyclic, non-transitive strategies and counter-strategies because of the combinatorial complexity and the dynamic relationships among agents. To our knowledge, the first Neural MMO challenge in 2021 \footnote{\textcolor{blue}{\url{https://www.aicrowd.com/challenges/the-neural-mmo-challenge}}} is the first competition that supports this FFA mode. In the competition, 16 participants' policies are evaluated together on the same map to benchmark their robustness and generalization. Our competition also follows this setting.

To achieve accurate evaluation and get more participants involved, we set up two tracks: PvE and PvP. In the PvE track, submitted policies are confronted with different levels of preset AIs, which can be seen as a fixed environment. This PvE setting reduces the uncertainty of the evaluation process and helps participants identify potential improvements. In the PvP track, we adopt the FFA setting as it can better benchmark the policy's robustness and also provides a persistent incentive for participants to improve their policies. Our competition is the first RL competition with this dual-track system, which was well received by our participants.

\section{Competition Orchestration}
\subsection{Environment}

\subsubsection{Introduction to Neural MMO} 
Neural MMO is an open-source research platform that simulates populations of agents in procedurally generated virtual worlds. It is inspired by classic massively multiagent online role-playing games (MMORPGs or MMOs for short) as settings where lots of players using entirely different strategies interact in interesting ways. Unlike other game genres typically used in research, MMOs simulate persistent worlds that support rich player interactions and a wider variety of progression strategies. We refer the reader to the original publication \cite{nmmo21} for full information on Neural MMO and its objectives. Our environment is adapted from version 1.5 of Neural MMO.

\subsubsection{Competition Configuration } 
The competition configuration of Neural MMO places $128$ agents in procedurally generated maps. Each map is 128x128 tiles. Scripted non-playable characters (NPCs) are spawned across the map. Agents must collect resources, \textit{Food} and \textit{Water} to survive and can attack each other and NPCs using three combat styles with strategic tradeoffs: \textit{Melee}, \textit{Mage}, and \textit{Range}. The competition focuses on robustness to new maps and new opponents and the team design introduces cooperation and specialization to different roles on top of this.

\subsubsection{Environment Wrapper}
To make the environment work with different agents, i.e., rule-based and RL agents, we wrap Neural MMO with two major changes. First, agents are spawned uniformly at the edges of the map. We randomize both the map seed and the initial position of each team across episodes to ensure a fair evaluation. Second, the observations of 8 agents in a team are grouped and made available to a single policy when they make decisions. A more strict setting in multi-agent cooperation might require that each agent compute its actions independently from teammates. We loosen that limit as in OpenAI Five \cite{five} and AlphaStar\cite{alphastar} in favor of enabling higher overall policy quality.

\begin{figure*}[t]
     \centering
     \begin{subfigure}[b]{0.32\textwidth}
         \centering
         \includegraphics[width=\textwidth]{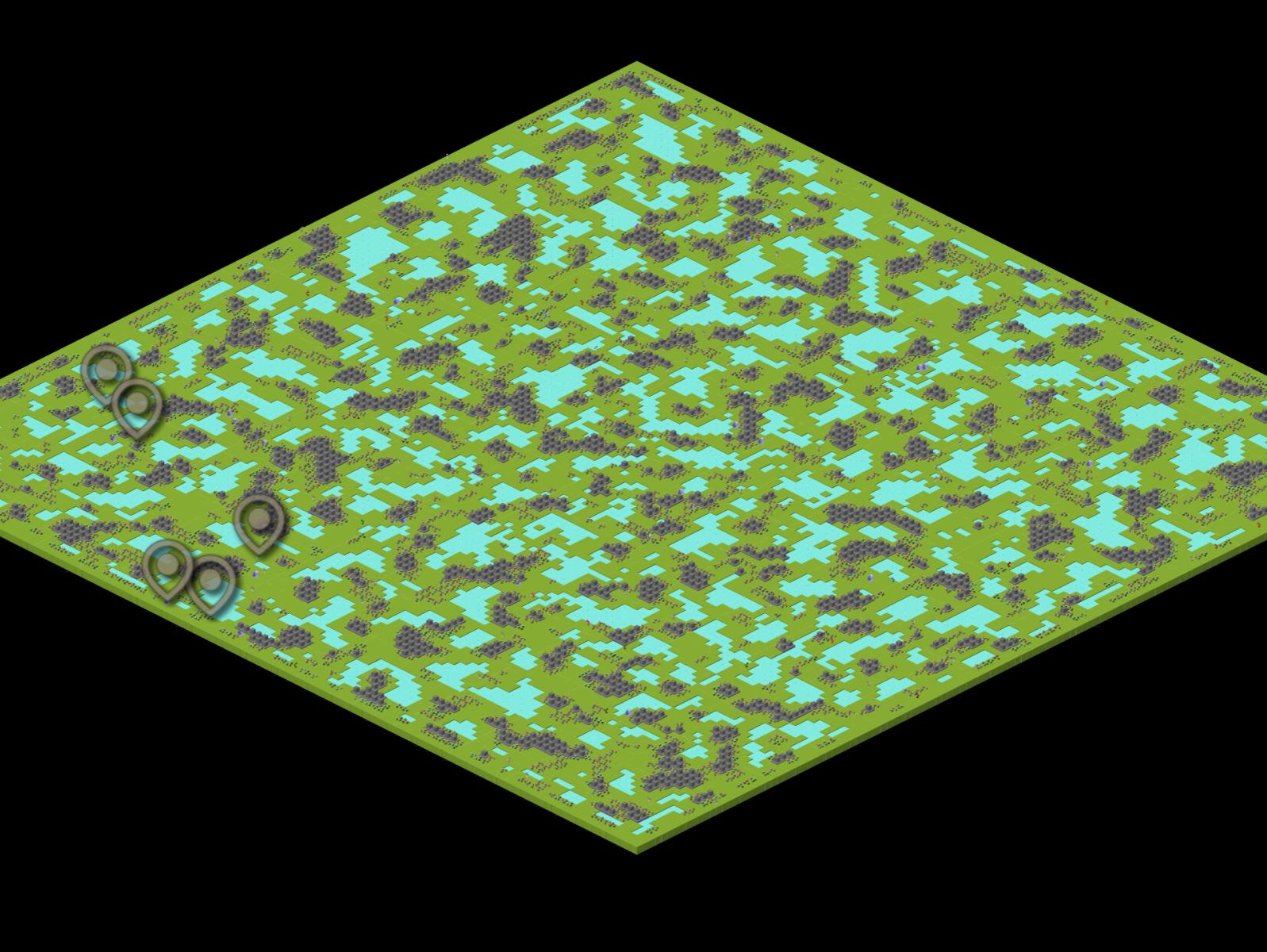}
         \caption{Overall: shows overall team position and resource distribution.}
         \label{fig:web-viewer-overall}
     \end{subfigure}
     \hfill
     \begin{subfigure}[b]{0.32\textwidth}
         \centering
         \includegraphics[width=\textwidth]{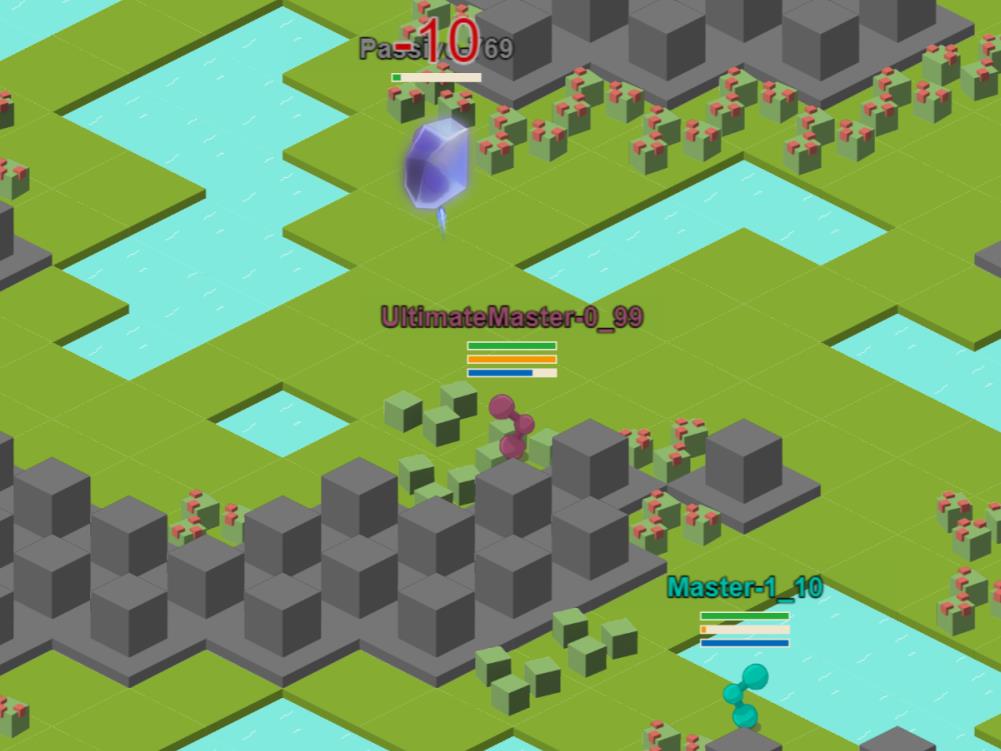}
         \caption{Close-up: shows important local details such as individual fights.}
         \label{fig:web-viewer-close-up}
     \end{subfigure}
     \hfill
     \begin{subfigure}[b]{0.32\textwidth}
         \centering
         \includegraphics[width=\textwidth]{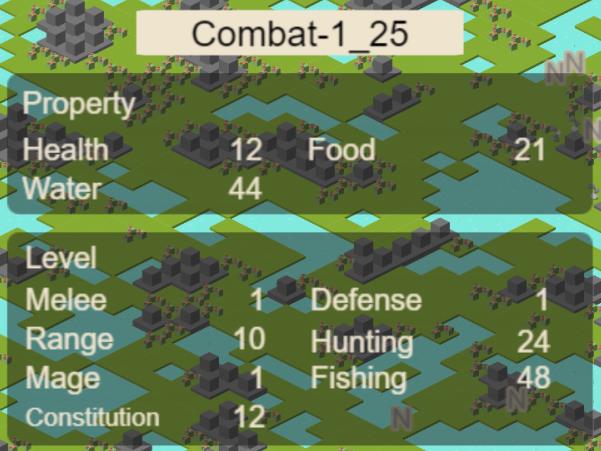}
         \caption{Details: shows the current numeric properties of the chosen agent.}
         \label{fig:web-viewer-details}
     \end{subfigure}
\caption{Web Viewer: a light visualization tool to show episode replays. (a) (b) (c) are three levels of view that can be altered during the playback. Users can rewind, pause, and change the playback speed on this web page. This allows participants to better understand the game and thus debug. }
\label{webviewer}  
\end{figure*}

\subsection{Resources}
For the participants' convenience, we have created a number of resources:
\begin{itemize}
    \item \textbf{Starter Kit}\footnote{\textcolor{blue}{\url{https://gitlab.aicrowd.com/neural-mmo/ijcai2022-nmmo-starter-kit}}}: a project containing all required segments to make a successful submission. With this guidance, new participants can make their first submission within 15 minutes.
    \item \textbf{Baseline}\footnote{\textcolor{blue}{\url{https://gitlab.aicrowd.com/neural-mmo/ijcai2022-nmmo-baselines}}}: an RL baseline implementation in a single file based on \textit{TorchBeast} \cite{torchbeast}. This provides RL researchers with a fundamental baseline to start with.
    \item \textbf{Env Docs}\footnote{\textcolor{blue}{\url{https://neuralmmo.github.io/build/html/rst/landing.html}}}: Documents and tutorials to help participants to get familiar with Neural MMO. 
    \item \textbf{Web Viewer}\footnote{\textcolor{blue}{\url{https://ijcai2022-viewer.nmmo.org/}}}: A light web replay viewer for our challenge, which allows participants with visual straightforward feedback for their policy development.  
\end{itemize}

\subsubsection{Web Viewer}
The web viewer is a light visualization tool to show the replays of the episodes, allowing our participants to review their policy's performance. For RL researchers, an accessible viewer is crucial to analyze learned strategies and improve their policies. The web viewer UI contains three levels of view:
\begin{enumerate}
    \item An overall view, as shown in Fig.\ref{fig:web-viewer-overall}, demonstrating the whole team's trajectories and the resource distribution of the global map;
    \item A close-up view, as shown in Fig. \ref{fig:web-viewer-close-up}, which reveals local details such as individual fights, including the attack target and attack style of each agent;
    \item A view of numeric details, as shown in Fig. \ref{fig:web-viewer-details}, which shows the current numeric properties of the chosen agent, such as its skill levels, current health, and collected resources.
\end{enumerate}

The right side of the interface shows the achievement scores of each team, allowing participants to interpret the varying abilities of each team to complete the four subtasks.

\begin{figure*}
    \centering
    \includegraphics[width=\linewidth]{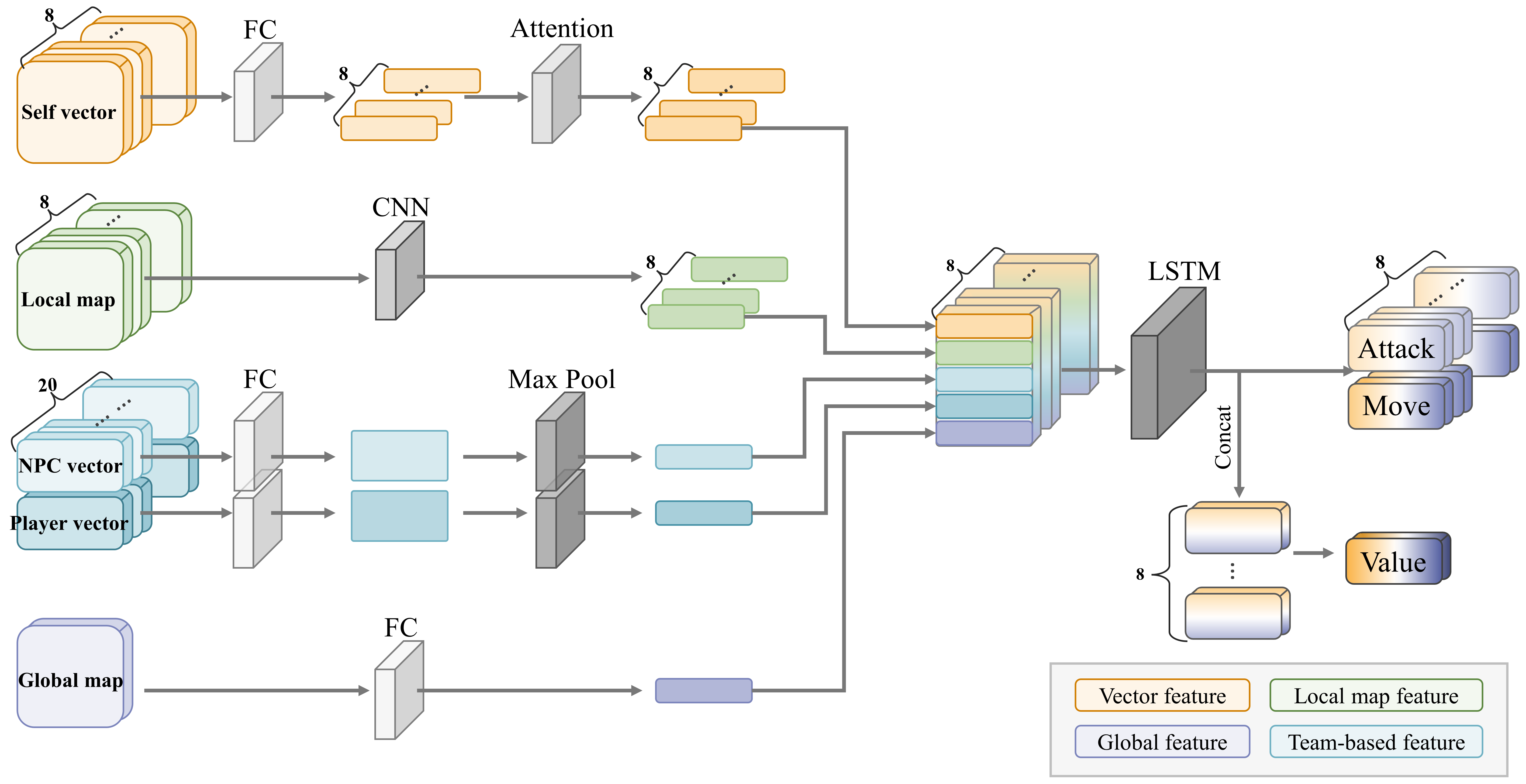}
    \caption{Model architecture for the PvE stage 3 baseline. It includes a core LSTM, a domain-specific observation encoder with sub-networks for flat, map, and set data, and a domain-specific decoder for fixed and variable-length actions.}
    \label{fig:network struture}
\end{figure*}

\subsection{Competition Structure}
The competition consists of two tracks: the PvE track and the PvP track. For clarity, PvE refers to one participant's policy vs. 15 built-in policies provided by the organizers. The PvE track serves as a fixed reference to help participants develop their policies. The PvE track contains 3 stages with different built-in policies and increasing difficulties.  In the main PvP track, 16 participants' policies are thrown into shared environments. This can better test a policy's robustness and generalization to opponents not seen during training.

\subsection{PvE (vs. fixed baselines)}

The main PvP track enables players to test the robustness and generalization of their agents against a variety of foes. However, there is a high degree of uncertainty: the quality of opponents changes over time, and the only measure of policy performance is relative to that of all other submissions. We have thus set up an additional PvE track of 3 stages with 3 main purposes:
(1) To help participants identify their agents' current performance against a fixed set of opponents of increasing quality;
(2) To further incentivize participation by providing three reasonably achievable milestones;
(3) To help participants understand the environment as guidance from easy to hard.

\begin{table*}[t]
    \centering
    \caption{Tasks defined in this competition to measure the performance of policies. Teams earn 0, 4, 10, or 21 points per task, depending on the hardest difficulty of that task completed by at least one agent in the team. \textit{Achievement} is defined as the sum of this score over all four tasks.}
    \label{tab:tasks}
    \begin{tabular}{llll} 
     \toprule
     Task & Easy(4 points) & Medium(10 points) & Hard(21 points) \\
     \midrule
     Travel the Lands & Explore 32 Meters & Explore 64 Meters & Explore 127 Meters \\ 
     Forage for Resources & Attain Skill Lvl 20 & Attain Skill Lvl 35 & Attain Skill Lvl 50 \\ 
     Secure an Advantage  & Acquire Lvl 1 Equipment & Acquire Lvl 10 Equipment & Acquire Lvl 20 Equipment \\ 
     Eliminate the Competition & Defeat 1 Player & Defeat 3 Players & Defeat 6 Players \\ 
     \bottomrule
    \end{tabular}
\end{table*}

\subsubsection{PvE Stage 1 (vs. Rule-Based Scripted Baselines)}
PvE stage 1 is a start-up stage arranged to help our participants familiarize themselves with the environment. We use three rule-based AIs named \textit{Combat}, \textit{Forage}, and \textit{Random}. The \textit{Combat} policy is hostile and will attack nearby agents. The \textit{Forage} policy focuses only on collecting resources and will attempt to flee from combat. The \textit{Random} policy moves randomly and is intended as a basic sanity check. Considering that all of these policies are open-source, this stage is intended to be relatively easy to beat.

\subsubsection{PvE Stage 2 (vs. RL Baselines)}
In PvE stage 2, we trained agents with PPO \cite{schulman2017proximal} using well-designed features extracted from raw observations. It is worth noting that we applied a \textit{decentralized training} method, which means each agent in our team can only get its own observations and act individually. This provides a medium-level reference for participants, which is much harder than that in Stage 1 but only takes RL agents about one week to conquer. 

\subsubsection{PvE Stage 3 (vs. Team-Based RL Baselines)}
The PvE stage 3 AI has the highest performance over all the 4 sub-tasks and can prevail over the PvE stage 2 baseline. The key distinction here is that we adopt a team-based method to process the whole team's observation and compute actions jointly. To be more specific, we devised a \textit{centralized training} strategy in which one network would concurrently process all 8 agents' observations and output the actions of eight agents. A team-based achievement is used as a reward. The advantage of this approach is that information is shared explicitly among teammates. Details of our modeling solution are as follows.

\textbf{Feature Design.} At the current scale, specialized feature extraction yields a large performance increase over processing raw observations. We divide, featurize, and process observations from all 8 agents on the team as follows:
\begin{itemize}
    \item \textbf{Member features}: Each team member's self-information such as their IDs (to better correlate to the index of the following local map feature), the initial positions (to measure the derivation of the current position), etc.
    \item \textbf{Enemy \& NPC features}: The observed entities' key features such as HP, level, and types are embedded here. This information helps agents learn how to behave in the presence of potential adversaries. We aggregate all 8 agents' observed entities together and use an additional vector of each entity to identify which agent is observing this entity to encourage the team-based policy to attack cooperatively;
    \item \textbf{Local map features}: Spatial information, such as the resource distribution on the observed local map, is embedded here to help agents learn to pathfind;
    \item \textbf{Global features:} Key global information such as time elapsed or the number of still-living teammates.
\end{itemize}

\textbf{Policy Architecture and Training:} The policy has three subnetworks: an observation encoder, the main long short-term memory (LSTM) network, and the action decoder. This architecture is shown in Fig.\ref{fig:network struture}. The input network contains fully-connected (FC) layers and max-pooling to process all scalar features and global information. Convolutional networks are used to process spatial information and attention modules are used to process position-invariant entity data. The main LSTM network processes the aggregate output of all of these encoders. The action output layers are normal FC layers. The policy is trained in a self-play setup against 15 teams controlled by the same policy. We employ valid action masks to accelerate exploration.

\textbf{Reward Design:} The competition scores teams based on their combat, foraging, and exploration. This mechanism is described in Section \ref{sec:evaluation} below; the relevant aspect here is that we use this function directly in order to compute rewards. 

\subsection{PvP (vs. other participants)}
Participants must pass the qualifying PvE Stage 1 in order to compete in PvP. Unlike in the PvE stages where opponents are fixed, the PvP opponent pool is dynamically sampled from the latest qualifying submissions from other participants. This includes reinforcement learned, scripted, and hybrid submissions.

We saw a large number of strategies emerge throughout the PvP stages, increasing our confidence in the platform as a proving ground for multi-agent reinforcement learning research, especially in testing the robustness of algorithms to new maps and opponents. We've noticed that some participants can rank high on PvE stage 1 or stage 2 but cannot maintain their advantage on the PvP stage, which indicates overfitting to the training domain.

\section{Evaluation System}
\label{sec:evaluation}

\begin{table*}[t]
\centering
\caption{Comparison of major RL competitions on the AICrowd platform, the primary venue for these events. Our competition has the most unique submitters and the highest sign-up-to-submission conversion rate.}
\label{tab_1}
\begin{tabular}{lcccc}
\toprule
\textbf{Competition} & \textbf{Views} & \textbf{Users} & \textbf{Submitted At Least Once} & \textbf{True Entry Rate} \\
\midrule
IJCAI 2022: Neural MMO Competition\footnotemark[8] & 40.3k & 540 & \textbf{111} & \textbf{20.50\%} \\

\hline
NeurIPS 2021: MineRL BASALT Competition\footnotemark[9] & 38.8k & 353 & 17 & 4.0\% \\
NeurIPS 2021: MineRL Diamond Competition\footnotemark[10] & 35.8k & 511 & 65 & 12.7\% \\
NeurIPS 2019: MineRL Competition\footnotemark[11] & 69.3k & 1124 & 41  & 3.6\% \\
\hline 
NeurIPS 2021 - The NetHack Challenge\footnotemark[12] & 49.1k & 584 & 46 & 7.9\% \\
NeurIPS 2020 Procgen Competition\footnotemark[13] & 52.8k & 711 & 85 & 12.0\% \\ 
\hline
Flatland 3\footnotemark[14] & 19.2k & 328 & 24 & 7.3\% \\
Flatland\footnotemark[15] & 72k & 1090 & 65 & 6.0\% \\
\hline
Unity Obstacle Tower Challenge\footnotemark[16] & 74k & 637 & 95 & 14.9\% \\
NeurIPS 2019: Learn to Move - Walk Around\footnotemark[17] & 37.9k & 364 & 71 & 19.5\%  \\
NeurIPS 2021 AWS DeepRacer AI Driving Olympics Challenge\footnotemark[18] & 20.4k & 337 & 41 & 12.2\% \\
Learn-to-Race: Autonomous Racing Virtual Challenge\footnotemark[19] & 24.1k & 476 & 53 & 11.1\% \\
\bottomrule
\end{tabular}
\end{table*}

Each participant's policy will control a team of 8 agents and will be evaluated in a free-for-all against 15 other teams on 128x128 maps. After 1024 environment steps, the team with the highest \textit{Achievement} wins.

\vspace{-2.5mm}
\subsection{Metrics}
\subsubsection{Multi-Task Metrics Definition}
To evaluate the generalization of the policy, we design a suite of 4 tasks, as shown in Table \ref{tab:tasks}. Each task has 3 difficulty levels: 4 points for easy, 10 points for normal, and 21 points for hard. Points were only awarded for the highest tier task completed in each category. The team with the most points at the end of a game (1024 steps) wins. This is a multi-objective task intended to be completed as a team: to achieve the maximum score for a task, only one agent on the team needs to complete it. This means it is reasonable for different agents on the team to employ different strategies. Such design encourages cooperation as a team and specialization of individual players.

\subsubsection{TrueSkill in PvP}
For the PvP track evaluation procedure,
we randomly select 16 submissions from all qualified submissions to start a PvP match. In the final evaluation, each submission will participate in approximately 1000 matches. The mean achievement score is not a good evaluation metric due to the variability of opponents. For example, model A gets a high score against weaker opponents and a low score against stronger opponents, while model B gets an above-average score against all levels of opponents. In this case, the mean achievement scores of the two models may be close, but it is obvious that model B is more robust. To more accurately measure the relative strength of the models, we use TrueSkill \cite{herbrich2006trueskill} to compute scores for each submission.

\subsubsection{Top 1 Ratio in PvE}
In the PvE track, the participant's model will play 10 matches against our built-in AI. With 16 teams per match, the variance of the mean achievement score for 10 matches is high. To evaluate the robustness of the policy, we use Top1Ratio as the evaluation metric. The Top1Ratio is the ratio of games won (i.e. highest score among all teams) over 10 matches. A Top1Ratio close to 1.0 indicates that the model is significantly stronger than the 15 built-in teams. 

\footnotetext[8]{\textcolor{blue}{\url{https://www.aicrowd.com/challenges/ijcai-2022-the-neural-mmo-challenge}}}
\footnotetext[9]{\textcolor{blue}{\url{https://www.aicrowd.com/challenges/neurips-2022-minerl-basalt-competition}}}
\footnotetext[10]{\textcolor{blue}{\url{https://www.aicrowd.com/challenges/neurips-2021-minerl-diamond-competition}}}
\footnotetext[11]{\textcolor{blue}{\url{https://www.aicrowd.com/challenges/neurips-2019-minerl-competition}}}
\footnotetext[12]{\textcolor{blue}{\url{https://www.aicrowd.com/challenges/neurips-2021-the-nethack-challenge}}}
\footnotetext[13]{\textcolor{blue}{\url{https://www.aicrowd.com/challenges/neurips-2020-procgen-competition}}}
\footnotetext[14]{\textcolor{blue}{\url{https://www.aicrowd.com/challenges/flatland-3}}}
\footnotetext[15]{\textcolor{blue}{\url{https://www.aicrowd.com/challenges/flatland}}}
\footnotetext[16]{\textcolor{blue}{\url{https://www.aicrowd.com/challenges/unity-obstacle-tower-challenge}}}
\footnotetext[17]{\textcolor{blue}{\url{https://www.aicrowd.com/challenges/neurips-2019-learn-to-move-walk-around}}}
\footnotetext[18]{\textcolor{blue}{\url{https://www.aicrowd.com/challenges/neurips-2021-aws-deepracer-ai-driving-olympics-challenge/}}}
\footnotetext[19]{\textcolor{blue}{\url{https://www.aicrowd.com/challenges/learn-to-race-autonomous-racing-virtual-challenge}}}

\subsection{Implementation}
We developed the distributed evaluation system shown in Fig. \ref{fig:framework} to quickly process submissions at scale. It can roll out hundreds of matches in parallel using k8s clusters and can return results within 10 minutes of submission.

We use the same distributed evaluation system for both the PvE and PvP tracks. The PvE track contains three levels; the main difference between them is the strength of the built-in AIs. Participants can enter the next level by reaching the specified Top1Ratio in the previous level. Reaching 25 points in PvE stage 1 qualifies a submission for the PvP track against other user submissions. Between PvE and PvP evaluation, we can accurately measure the strength of all models relative both to each other and to fixed baselines. The PvP evaluation is run once per week while PvE evaluation proceeds immediately upon submission. This ensures fast feedback to inform development at all times and more extensive feedback weekly.

\section{Summary and Analysis}

\subsection{Summary of the competition}
The competition received over 40k views, 537 individual signups, 110 team signups, and 1679 submissions. This makes it one of the largest RL competitions to date, outpacing all of the MineRL competitions thus far and Nethack -- despite having a significantly higher barrier to entry due to the complexity of the task, lack of a single-agent track, lack of offline data, and complex observation and action representation. Of these participants, 48 teams were able to pass our first-round qualifier. 20 teams were able to win at least some games versus better policies that we trained for round 2, with 16 qualifying for round 3. We trained much stronger baselines for these rounds, but 7 teams were still able to win at least some games, and 6 were convincingly better than our best baseline. The best policies fully accomplished the task of the competition. Table \ref{tab_1} compares the metrics of major RL competitions and demonstrates the scope of our contest.

\subsection{Analysis of the Competition Design}

\begin{figure}[t]
    \centering
    \includegraphics[width=\linewidth]{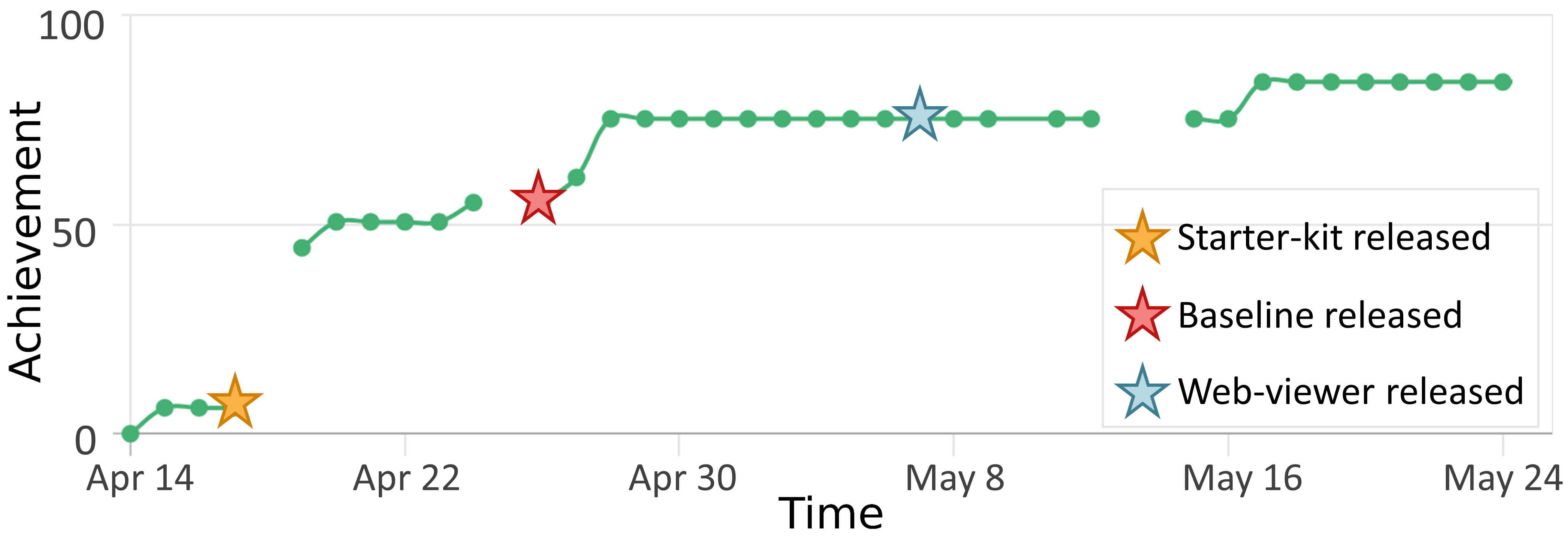}
    \caption{Maximum achievement in PvE stage 1 through time, measured over all participants. The release of the baseline corresponds with a large jump in submission quality.}
    \label{fig:Ach}
\end{figure}

\begin{figure}[t]
    \centering
    \includegraphics[width=\linewidth]{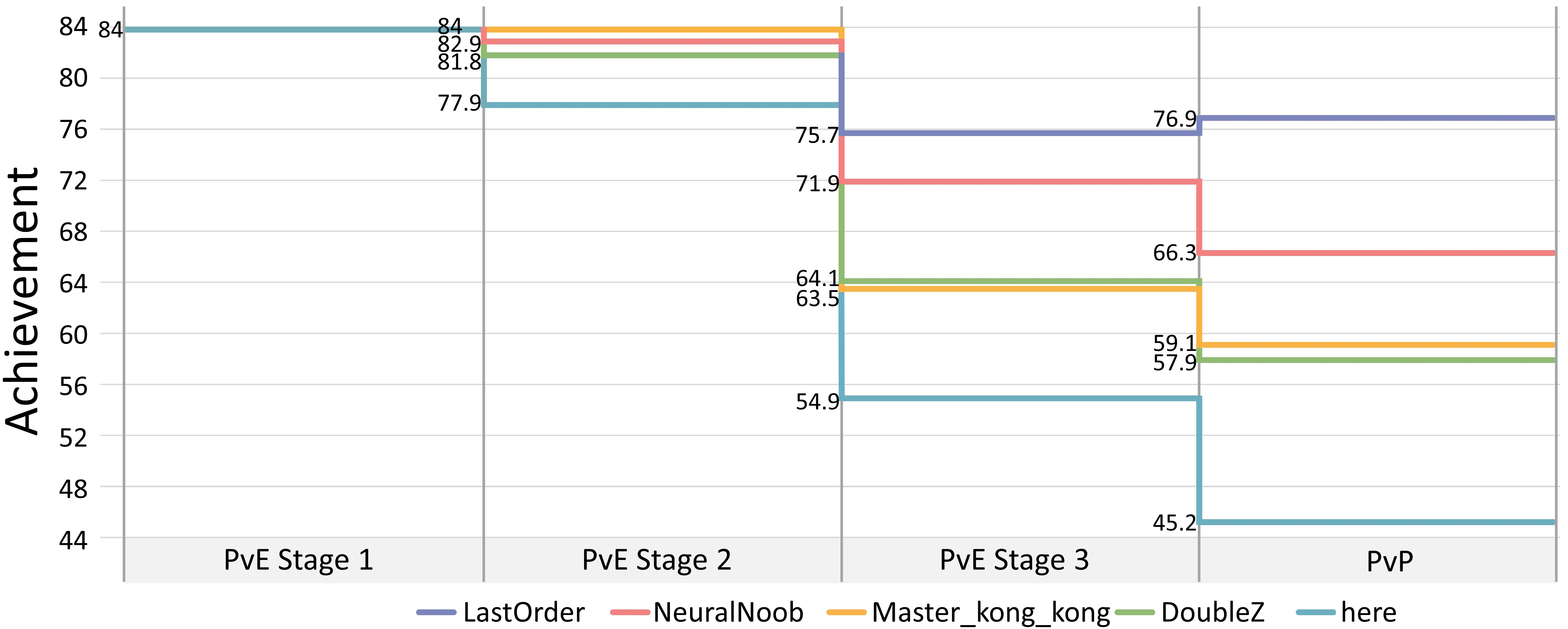}
    \caption{Top five participant scores in each round. The increasing difficulty of later PvE rounds corresponds with a decline in achievement score. Performance in stage 3 is comparable to performance in the last PvP stage.}
    \label{fig:Top_5_per}
\end{figure}

Fig.\ref{fig:Ach} shows the increases of max achievement over time. We can find that the three sudden rises are due to the starter kit release, the official baseline release, and the web viewer release: these tools were either useful or at least motivational to participants.

Fig.\ref{fig:Top_5_per} shows performance against different stages of the PvE track. The baseline quality increases across rounds, so submission performance declines as expected from stage to stage. Interestingly, the final PvE stage results are similar to those of the final PvP track. This suggests that this track was effective in allowing participants to quickly evaluate their submissions. This is useful because the PvP stage is more computationally expensive, so we can only run it once per week.

\subsection{Analysis of 1600+ Policies}
We gathered over 1600 submissions and categorized them as rule-based methods (behavior tree, planning-based methods, heuristic methods, etc.) or learning-based methods (reinforcement learning) based on the algorithms employed by the participants. \textbf{We additionally release presentations from the top 5 teams about their approach (camera ready for anonymity).}

\begin{figure}[t]
    \centering
    \includegraphics[width=\linewidth]{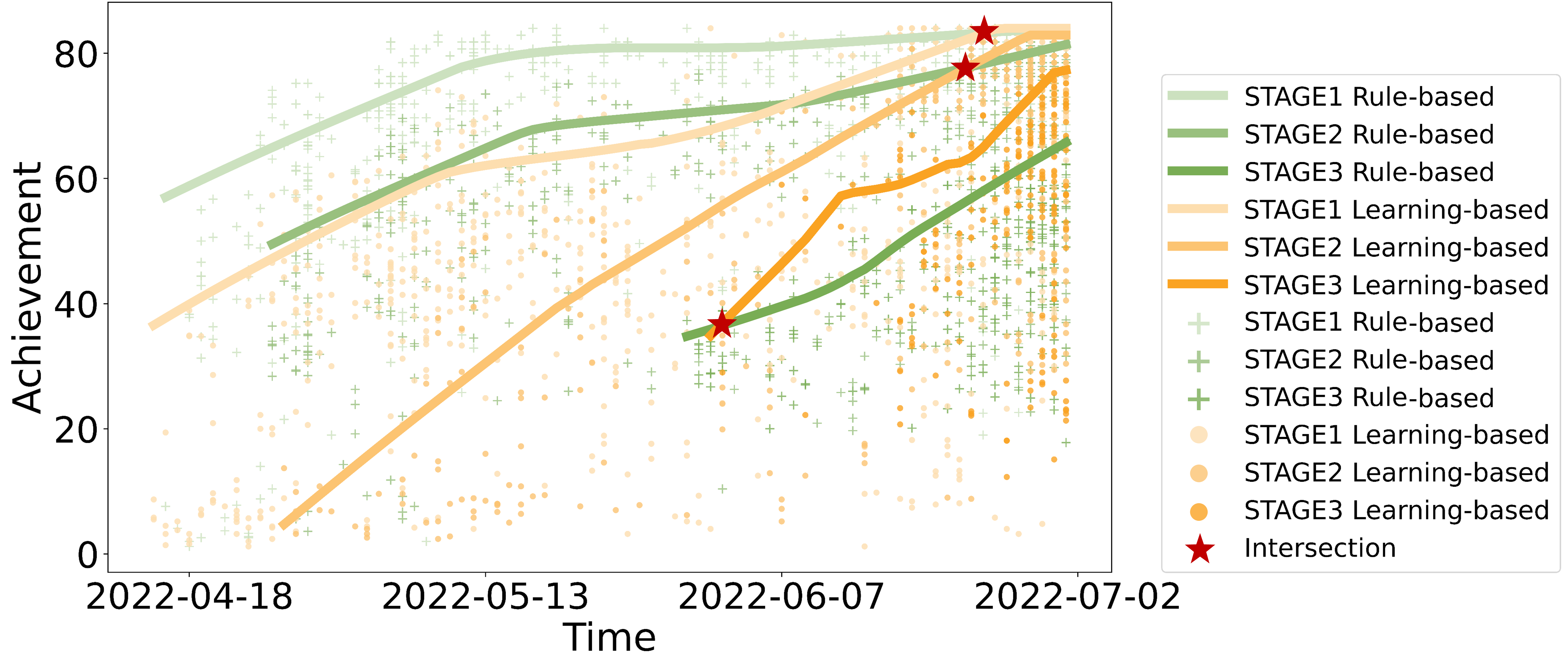}
    \caption{The effectiveness of Rule-Based and Learning-Based methods across three PvE stages. The six lines represent the peak performance of the approach at each stage. The red star marks the point at which the highest performance of the two approaches overlaps.}
    \label{fig:time_achievement_intersection}
\end{figure}

Fig.\ref{fig:time_achievement_intersection} illustrates the best achievements over time for both classes of methodologies in the three stages of PvE and PvP track. We make several observations. (1) Rule-based or learning-based methods both achieve satisfactory performance. (2) The performance curves of rule-based and learning methods cross earlier in later stages. This suggests that rule-based methods are quick to get working but do not scale as well against complex opponents. (3) The green lines of rule-based methods in Stage 1 and Stage 2 almost converge and still climb in Stage 3. The orange lines of learning-based methods are all climbing. Thus, there is still room for further research even on this version of Neural MMO, without even considering some of the more recent additions to the environment.

\subsubsection{Robustness and Generalization.} 
\begin{figure*}[htbp]
    \centering
    \includegraphics[width=1.0\textwidth]{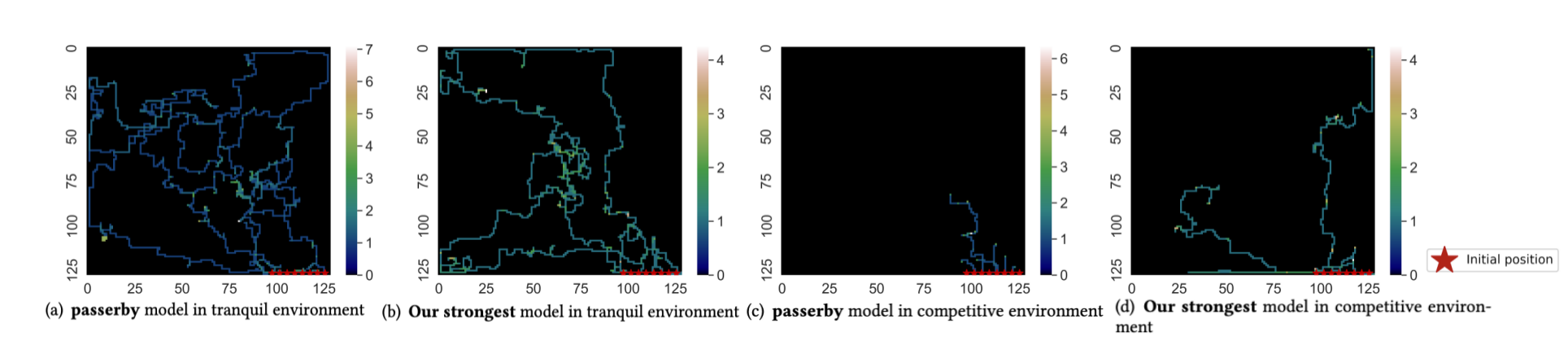}
    \caption{The figure depicts the movement trajectory of the participant model against weak (PvE stage 1) and strong (PvE stage 3) opponents. We have compared our strongest model in the same context. \textbf{Passerby} employs different pathing strategies against different opponents, demonstrating that our competition may be utilized to evaluate policy robustness.}
    \label{passby}
\end{figure*}

As seen in Fig.\ref{fig:Top_5_per}, the performance of the participants' models varies at different stages. As an example, plotted the exploration pattern of the winning policy \textit{passersby}, against a weaker opponent (PvE stage 2) and against a stronger opponent (PvE stage 3). This player's model explores significantly more against inferior opponents than against stronger ones, which is shown in Fig.\ref{passby}. This further demonstrates that this environment may facilitate the study of model robustness and generalization by introducing diverse adversaries.

\begin{figure*}[htbp]
    \centering
    \includegraphics[width=\linewidth]{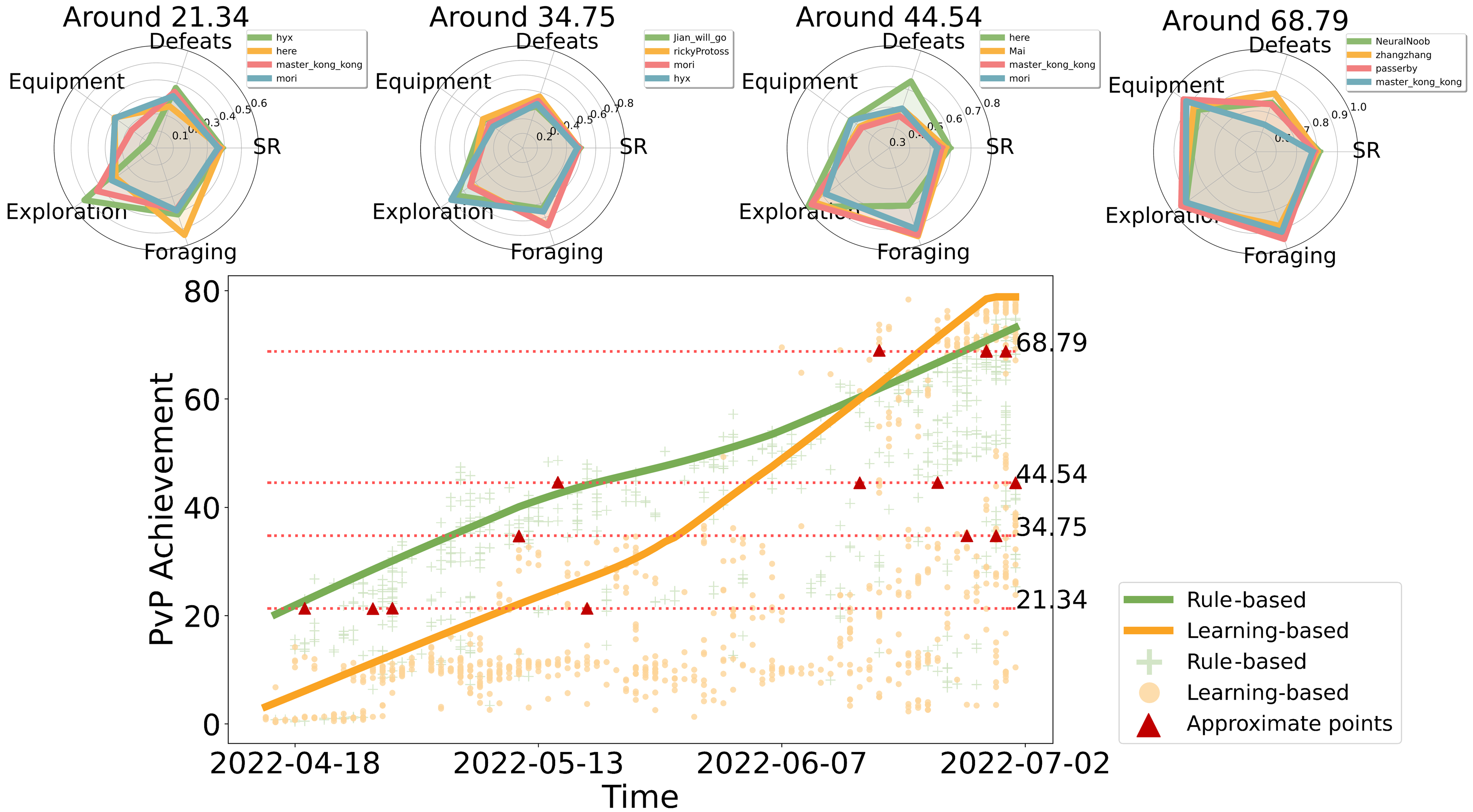}
    \caption{We compared the performance of players with the same achievement on each of the four subtasks. Even if the final achievements are identical, participants will employ different methods to accomplish the job, demonstrating that Neural MMO can accommodate a variety of tactics.}
    \label{fig:radar}
\end{figure*}

\subsubsection{Diversity of Policies.} We find that policies that achieve the same score may employ different strategies, as indicated by differing performance on the four subtasks and the overall trajectories of different agents. The models of different players can have varied strengths and weaknesses on the sub-tasks as shown in Fig.\ref{fig:radar}. Using the four players with final achievements close to 68.79 as an example, \textit{zhangzhang}'s model achievement on \textit{Defeat} is high, but their \textit{Equipment} is inadequate, indicating that their agents' primary tactic is to kill other players‘ agents. The superior performance of \textit{kongkong}'s agents in both \textit{Foraging} and \textit{Equipment} suggests that their strategy is more adept at utilizing map resources. 

\begin{figure*}[htbp]
    \centering
    \includegraphics[width=\linewidth]{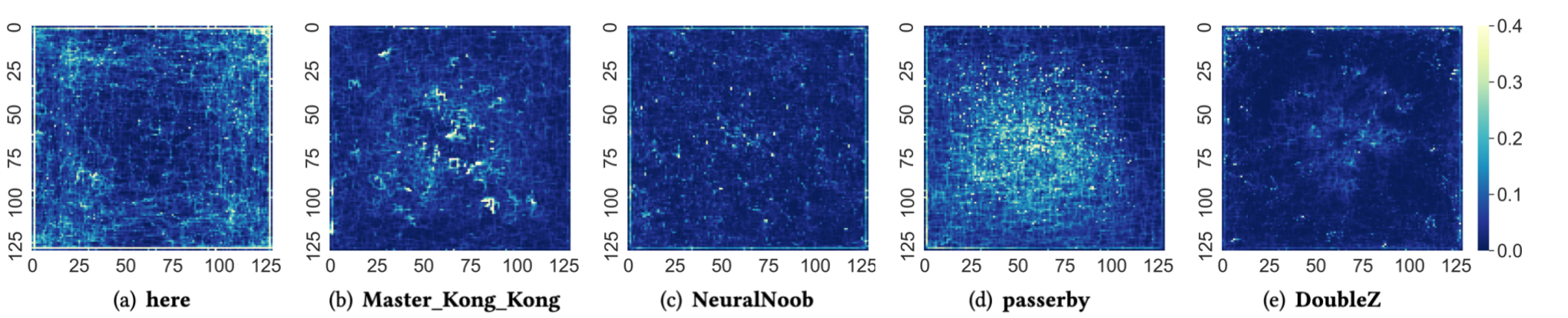}
\caption{Visitation frequency of various policies, computed by summing per-tile exploration counts over over 50 episodes. Different policies demonstrate distinct exploration preferences. For example, the \textit{here} policy primarily explores around the edges of the map while \textit{passerby} spends more time in the center of the map.}    
\label{heatmap}    
\end{figure*}

\begin{figure*}[htbp]
    \centering
    \includegraphics[width=1.0\textwidth]{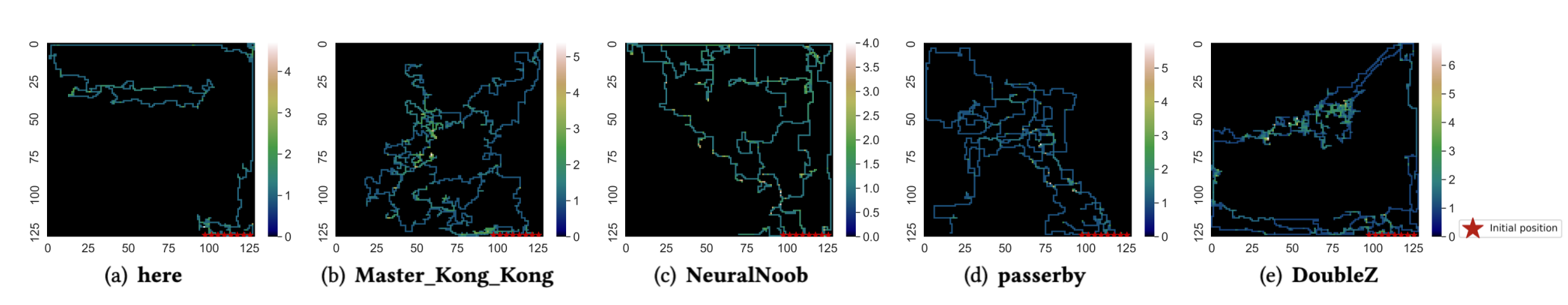}
    \caption{Movement path of five teams. Different policies employ different pathing strategies, with some choosing to disperse at the start and converge at the center while others explore as a team.}
\label{trajectory}    
\end{figure*}

For the trajectories shown in Fig.\ref{trajectory}, we choose the paths of the top five ranking submissions and observe that various teams have distinct navigation preferences. The team \textit{here}, for instance, will explore in a straight line to maximize their exploration score, whereas the team \textit{DoubleZ} will go deeper into the heart of the map from the beginning because there are higher-level NPCs there, allowing them to quickly upgrade their equipment. \textit{Master\_kong\_kong}'s team will do the most comprehensive exploration, allowing them to become familiar with the entire area more quickly. Similarly, we count the frequency of visits to each tile by the agents under different strategies, and we can find that there will be differences among models as shown in Fig.\ref{heatmap}.

\section{Conclusion}

To benchmark the robustness and generalization of MARL algorithms, we hosted a multi-agent artificial intelligence challenge and received 1600+ policy submissions. The top five submissions all surpassed the best existing baselines while employing strategies ranging from rule-based to full RL. However, the performance curve till the end of the competition indicates that policies have not yet reached the performance upper bound in the environment and that there is still considerable potential for RL algorithms in further research.

From an algorithmic perspective, the results of this competition and the analysis of the top-ranking solutions demonstrate that the conceptually simple methods effective in large-scale industry research can also work on complex but academic-scale tasks. We suggest that a gap in tooling and infrastructure, rather than purely algorithms, is the main short-term bottleneck preventing reinforcement learning from working on complex, multi-agent environments. We argue that the simplest way to realize this result in other environments is to run competitions and open-source the tools built by organizers and participants. By aggregating these implementations across multiple domains, we may begin to see the commonalities and build more general-purpose tools. We hope that our work will inspire others to adopt the competition model of research and open-source their tooling as we have.

\bibliographystyle{ACM-Reference-Format}
\bibliography{sample}
\end{document}